# A Text Classification-Based Approach for Evaluating and Enhancing the Machine Interpretability of Building Codes

Zhe Zheng[1], Yu-Cheng Zhou[1], Ke-Yin Chen[1], Xin-Zheng Lu[1], Zhong-Tian She[1], Jia-Rui Lin[1,*]

Department of Civil Engineering, Tsinghua University, Beijing, 100084, China

*Corresponding author, E-mail: lin611@tsinghua.edu.cn, jiarui_lin@foxmail.com

**Abstract:**

Interpreting regulatory documents or building codes into computer-processable formats is essential for the intelligent design and construction of buildings and infrastructures. Although automated rule interpretation (ARI) methods have been investigated for years, most of them highly depend on the early and manual filtering of interpretable clauses from a building code. While few of them considered machine interpretability, which represents the potential to be transformed into a computer-processable format, from both clause- and document-level. Therefore, this research aims to propose a novel approach to automatically evaluate and enhance the machine interpretability of single clause and building codes. First, a few categories are introduced to classify each clause in a building code considering the requirements for rule interpretation, and a dataset is developed for model training. Then, an efficient text classification model is developed based on a pretrained domain-specific language model and transfer learning techniques. Finally, a quantitative evaluation method is proposed to assess the overall interpretability of building codes. Experiments show that the proposed text classification algorithm outperforms the existing CNN- or RNN-based methods, improving the F1-score from 72.16% to 93.60%. It is also illustrated that the proposed classification method can enhance downstream ARI methods with an improvement of 4%. Furthermore, analyzing the results of more than 150 building codes in China showed that their average interpretability is 34.40%, which implies that it is still hard to fully transform the entire regulatory document into computer-processable formats. It is also argued that the interpretability of building codes should be further improved both from the human side (considering certain constraints when writing building codes) and the machine side (developing more powerful algorithms, tools, etc.).

# 1 Introduction

The architecture, engineering, and construction (AEC) industry is undergoing a significant transformation from traditional labor-intensive methods to digital, automated, and smart methods (Wu et al., 2022; Liao et al., 2021). Building regulatory documents are used to guarantee the safety, sustainability, and comfort of the entire lifecycle of a built environment. Thus, interpreting regulatory documents or building codes into computer-processable formats is essential for the intelligent design and construction of buildings and infrastructures (Wu et al., 2022a). However, the existing codes are still written in natural language and mainly read and utilized by domain experts, which is difficult for computers to automatically understand, process, and analyze (Ismail et al., 2017). Therefore, automated rule interpretation methods which aim to automatically interpret the regulatory texts into a computer-processable format have been studied by many researchers for automated compliance checking and even intelligent design (Eastman et al., 2009; Ismail et al., 2017; Sobhkhiz et al., 2021; Fuchs, 2021).

Many studies have focused on automated rule interpretation tasks based on natural language processing (NLP) (Fuchs, 2021). For example, information extraction and transformation have been widely studied by many researchers (Zhang & El-Gohary, 2015; Zhou et al., 2022; Zhang & El-Gohary, 2021a). Information extraction and transformation first identify and extract the words and phrases in the relevant sentences and then interpret the extracted semantic information elements into computer-processable representations. However, the building code is usually written in natural language and used by human engineers with extensive domain knowledge, problems of incompleteness, ambiguity, vagueness, and tacit knowledge have made it difficult for a computer to understand some complex building code automatically (Soliman-Junior et al., 2021). Not all requirements in the codes could be interpreted automatically by a computer. Regarding these requirements, the interpretability is relatively low. And the requirements with low interpretability can lead to a large number of errors if interpreted automatically. Therefore, although existing studies have facilitated the process of automated rule interpretation, information extraction and transformation efforts mainly focus on the early and manually filtered clauses that can be easily interpreted. In recent years, to further improve the accuracy of the information extraction and transformation processes, text classification has been incorporated into the automated rule interpretation pipeline. Text classification aims to recognize relevant sentences in a regulatory text corpus, thereby preventing inefficiency and errors in downstream tasks, which result from the unnecessary processing of irrelevant text (Zhou & El-Gohary, 2016; Song et al., 2018). However, the existing text classification efforts mainly focus on classifying clauses according to their topics, subjects, or scopes (Zhang & El-Gohary, 2021b). However, most of them neglect the interpretability of the clause, which reflects its potential to be transformed into a computer-processable format.

Knowledge of the types and interpretability of clauses is essential for improving or extending existing ARC systems to achieve enhanced coverage and performance or to develop new ARC systems with increased capabilities (Solihin & Eastman, 2015; Zhang & El-Gohary, 2021b; Soliman-Junior et al., 2021). In addition, the evaluation of the interpretability of an entire building code may also lead to the optimization and revision of the existing building codes, making them more suitable for ARC systems (Soliman-Junior et al., 2021). However, limited research efforts have been devoted to identifying and characterizing the different types of clauses in AEC regulations to better assess the interpretability of all clauses (Solihin & Eastman, 2015). For example, Solihin & Eastman categorized clauses into four categories based on what type of building information modeling (BIM) data the clauses required (Solihin & Eastman, 2015). However, they evaluated interpretability of each clause through a manual method, which was labor intensive and time consuming. It is difficult to evaluate the interpretability of a large number of existing clauses and building codes. Zhang & El-Gohary automatically evaluated the interpretability of the International Building Code (IBC) and its variations/amendments based on a clustering approach, thereby improving the efficiency of the evaluation process (Zhang & El-Gohary, 2021b). However, their clustering-based method has only been applied to the IBC thus far, and the method's applicability to other building codes needs to be further analyzed. The existing studies mainly analyze the interpretability of a single clause based on manual or semi-automated means. Few works analyze the interpretability of the whole building codes for the optimization and revision of the existing building codes. Therefore, the research gaps include: a) automated interpretability evaluation methods for both single clause and building codes, b) quantitative analysis of the influence of rule interpretability evaluation on automated rule interpretation, and c) interpretability evaluation of existing large-scale building codes for future code development, optimization, and revision.

To better evaluate the interpretability of existing building codes, this work proposes an automated interpretability evaluation method based on text classification. First, category criteria that are suitable for evaluating the interpretability of each clause are proposed, and subsequently, a training dataset is constructed based on the proposed category criteria. Second, a deep learning (DL) model that utilizes transfer learning techniques is trained to automatically classify each clause; this approach outperforms the widely used traditional DL models and achieves state-of-the-art results. Third, we proposed an interpretability evaluation method for a whole building code based on the well-trained DL model. Finally, the proposed method is applied to assess the overall interpretability of a series of Chinese building codes.

The remainder of this paper is organized as follows. Section 2 reviews the related work and highlights the potential research gaps. Section 3 describes the category criteria designed for interpretability evaluation, the development of the dataset, and the interpretability evaluation method for building codes. Section 4 describes the training process and performance evaluation of DL models. Section 5 conducts an experiment to demonstrate the improvement achieved by the proposed interpretability evaluation method in terms of rule interpretation. Section 6 evaluates the interpretability of a series of Chinese codes. Section 7 discusses the advantages and contributions of this research and notes its limitations. Finally, Section 8 concludes this research.

## 2 Overview of related studies

### 2.1 Evaluation of the interpretability of building codes

Automated rule interpretation, which aims to interpret regulatory clauses into a computer-processable format, is the most vital and complex stage of ARC. To achieve a full level of ARC, many automated rule interpretation attempts have been made based on NLP techniques (Fuchs, 2021). Existing studies can be broadly divided into two groups, including (1) information extraction and transformation studies, which extract relevant semantic elements from clauses and then format them into computer-processable representations (Zhang & El-Gohary, 2015; Zhang & El-Gohary, 2017; Zhou et al., 2022; Zhang & El-Gohary, 2021a), and (2) text classification research, which filters irrelevant clauses in a building code, thereby improving the accuracy of downstream tasks (e.g., information extraction) (Zhou & El-Gohary, 2016; Salama & El-Gohary, 2016; Song et al., 2018). Although existing studies have facilitated automated rule interpretation, information extraction and transformation efforts have mainly focused on filtered clauses that can be easily interpreted. These techniques cannot measure the interpretability of all types of clauses. In addition, text classification efforts have mainly focused on classifying clauses according to their topics, subjects, or scopes (Zhang & El-Gohary, 2021b), which makes it difficult to evaluate the interpretability of clauses.

Significant benefits are obtained if knowledge regarding clauses' types and interpretability can be measured. The potential benefits may include (1) the ability to reuse clause's structures, and best practices (Solihin & Eastman, 2015); (2) the improvement or extension of the existing ARC systems to achieve enhanced coverage and performance (Solihin & Eastman, 2015; Zhang & El-Gohary, 2021b; Soliman-Junior et al., 2021); and (3) the optimization and revision of the existing building codes to make them more suitable for ARC systems (Soliman-Junior et al., 2021).

However, limited research has been devoted to identifying and characterizing the different types of clauses in AEC regulations to better assess the interpretability of all clauses (Solihin & Eastman, 2015). The existing efforts in evaluating the interpretability of building codes are list in Table 1. Moreover, most of the existing efforts with respect to interpretability evaluation are based on manual methods, which makes it difficult to evaluate the interpretability of large-scale building codes. For example, Solihin & Eastman categorized clauses into four categories based on what types of BIM data the clauses required (Solihin & Eastman, 2015). Uhm et al. analyze the interpretability of requirements in requests for proposals (RFPs) for building designs in South Korea. The results show that only 14% of are computer-interpretable sentences (Uhm et al., 2015). Malsane et al. classified building regulation clauses into those that were computer-interpretable (declarative) and those that were not (informative) (Malsane et al., 2015). They found that 19.7% of the clauses in the England and Wales Building Regulations that relate to fire safety for dwelling houses are computer-interpretable (Malsane et al., 2015). Soliman-Junior et al. conducted a comprehensive assessment regarding the interpretability of UK healthcare building codes (Soliman-Junior et al., 2021). Their results show that 47% of clauses in Healthcare Design Regulations in the UK can be transformed into a logical rule sentence (Soliman-Junior et al., 2021). In recent years, with the development of machine learning (ML) techniques, some automated methods have emerged. Among the automated evaluation methods, Zhang & El-Gohary evaluated the interpretability of the IBC based on a clustering approach (Zhang & El-Gohary, 2021b). However, their clustering-based method has only been applied to the IBC thus far, and the method's applicability to other codes needs to be further analyzed. Their analysis results show that 32.6% of clauses in the International Building Code have high or moderately high computability (Zhang & El-Gohary, 2021b).

Table 1 shows that the interpretability of different codes in different countries is not the same. In order to promote the development of automated rule interpretation, it is necessary to analyze the interpretability level of codes in a country to facilitate the optimization and revision of the existing building codes and to improve the interpretability of future codes. Therefore, methods for automatically evaluating the interpretability of single clause and large-scale existing building codes should be further developed. Since many existing methods evaluate interpretability via manual text classification, a natural idea is to automate the process via DL and NLP methods.

Table 1 Existing efforts in evaluating the interpretability of building codes

| Reference | Code | Interpretability | Method |
| --- | --- | --- | --- |
| Solihin & Eastman, 2015 | Singapore Building Codes, Singapore Fire Code, International Building Code, et al | / | Manual |
| Uhm et al., 2015 | Requests for proposals for building designs in South Korea | 14% requirements are computer-interpretable | Manual |
| Malsane et al., 2015 | England and Wales Building Regulations that relate to fire safety for dwelling houses | 19.7% requirements are computer-interpretable | Manual |
| Soliman-Junior et al., 2021 | Healthcare Design Regulations in the UK | 47% requirements can be transformed into a logical rule sentence | Manual |
| Zhang & El-Gohary, 2021b | International Building Code | 32.6% requirements with high or moderately high computability | Automated |

## 2.2 Automated text classification methods in the AEC domain

The text classification task involves recognizing relevant sentences from large quantities of documents and assigning them to one or more predefined categories (Manning & Schutze, 1999) to facilitate downstream tasks. Many automated text classification methods have been proposed in the AEC domain, and we summarize them in Table 2.

Because ML and DL algorithms cannot directly process texts in natural language, they typically represent a sentence by numerical features (Hassan & Le, 2020). The widely used feature representation methods include (1) bag-of-words (BOW) models and (2) word embedding models. In a BOW model, each sentence is represented as a numeric vector, where each element in the vector corresponds to a word. Each value in the vector can either be zero, indicating the absence of a word in the sentence, or a real number, indicating the frequency of the word in the sentence (Salton & Buckley, 1988; Hassan & Le, 2020). Unlike BOW models, a word embedding model employs artificial neural networks to generate a multidimensional real number vector that represents the semantics of every unique word in the whole input corpus (Mikolov et al., 2013). The word embedding model assumes that words appearing in the same context may have similar meanings. Thus, the vectors of these similar words are brought closer together in the vector space (Mikolov et al., 2013; Hassan & Le, 2020).

After the features are represented, ML or DL methods are used to conduct the category prediction. The adopted methods can be broadly divided into three categories according to model complexity: shallow ML models, traditional DL models, and pretrained DL models. The shallow ML models, which are the simplest type, include naïve bayes (NB) models, support vector machines (SVMs), logistic regression (LR), and so on. With the development of DL techniques, some traditional DL models have emerged, including convolutional neural network (CNN)-based models (e.g., TextCNN (Liu et al., 2016) and deep pyramid convolutional neural networks (DPCNN) (Johnson et al., 2017)) and recurrent neural network (RNN)-based models (e.g., TextRNN (Lai et al., 2015)). This type of model can automatically learn data representations with a nonlinear combination of multiple processing layers based on training data (LeCun et al., 2015) and performs better than shallow models (Zhong et al., 2020; Cheng et al., 2020; Tian et al., 2021). The main drawback of traditional DL models is that they require highly expensive manual efforts to prepare sufficient training datasets (Xu & Cai, 2021). Therefore, in recent years, pretrained DL models (e.g., bidirectional encoder representation from transformers (BERT) (Devlin et al. 2018)) have been proposed. These models are pretrained to provide a useful initialization of the parameter weights so that new tasks can be learned from large datasets (Fang et al. 2020). Pretrained models can achieve better performance than traditional DL models when the training datasets are relatively small (Fang et al. 2020). Therefore, in recent years, BERT-based text classification methods are widely used in various domains. For example, Tagarelli & Simeri (Tagarelli & Simeri, 2022) proposed LamBERTa (Law article mining based on BERT architecture) for the law article retrieval tasks. Tian et al. (Tian et al., 2022) proposed a BERT and graph convolutional neural network-based safety-hazard classification method for large-scale construction projects. Wang et al. (Wang et al., 2022) proposed a BERT-based text classification for Chinese emergency management with a novel loss function.

However, the widely used pretrained models are pretrained on a general-domain corpus (Sun et al., 2019), which has a different data distribution than that of the AEC domain. When the given training datasets are very limited, the common pretrained models still struggle to obtain satisfactory results. Therefore, more studies should be conducted to further improve the performance of pretrained DL models.

Table 2 Automated text classification methods in the AEC domain

| Methods/Features | BOW | Word Embedding |
| --- | --- | --- |
| Shallow ML models | Caldas et al., 2002; Salama & El-Gohary, 2016; Zhou & El-Gohary, 2016; Hassan & Le, 2020 | Hassan & Le, 2020; Tian et al., 2021 |
| Traditional DL models | Hassan & Le, 2020 | Song et al., 2018; Cheng et al., 2020; Zhong et al., 2020; Fang et al., 2020; Tian et al., 2021 |
| Pretrained DL models |  | Fang et al., 2020 |

## 2.3 Research gaps

Despite the importance of these efforts, this paper aims to address four knowledge gaps.

First, few studies have focused on automated evaluation of machine interpretability from both clause-level and document-level. Most of the existing studies focused on automated rule interpretation for filter irrelevant clauses at the sentence level. The existing research regarding the evaluation of clause interpretability is based on manual methods, which makes it impossible to quickly evaluate the interpretability of a large number of existing building codes.

Second, due to the lack of automated evaluation methods, few works have quantitatively analyzed the influence of rule interpretability evaluation on automated rule interpretation.

Third, because of the lack of automated evaluation methods, few works have conducted systematic analyses and evaluations regarding the interpretability of the existing building codes. Therefore, it is difficult to optimize and revise the existing building codes to make them more suitable for ARC systems.

# 3 Methodology

To address the abovementioned problems, this article proposes a text classification-based approach to automatically evaluate the interpretability of every single clause and building code. The workflow of this method is shown in Fig. 1. It consists of three parts, including part 1: classification model, part 2: clause-level interpretability evaluation, and part 3: document-level interpretability evaluation.

In part 1, we first propose clause-level category criteria, which aim to measure the interpretability of clauses (Section 3.1). Subsequently, based on the category criteria, a dataset for model training is constructed via manual annotation (Section 3.2). Then, we adopt a well-known pretrained DL model (i.e., BERT) for training and prediction. In addition, to improve the model's performance in a case with small samples, this work adopts a domain corpus to further enhance the pretrained DL model (Section 3.4). A series of models and experiments are utilized for comparison and evaluation purposes (Section 4). The input of this part is clauses, and the outputs include an open-source dataset and a well-trained DL model.

In part 2, the well-trained DL model and the proposed clause-level interpretability evaluation method are integrated into an automated rule interpretation method, to enhance the performance of the automated interpretation (Section 5). The input of this part is clauses, and the output is the enhanced automated rule interpretation pipeline.

In part 3, based on the clause-level category criteria, a quantitative indicator for evaluating the document-level interpretability of building codes is proposed (Section 3.3). Associated with the well-trained DL model, we perform the interpretability evaluation analysis for more than 150 building codes in China (Section 6). The input of this part is a list of building codes, and the outputs are suggestions for code development (Section 7).

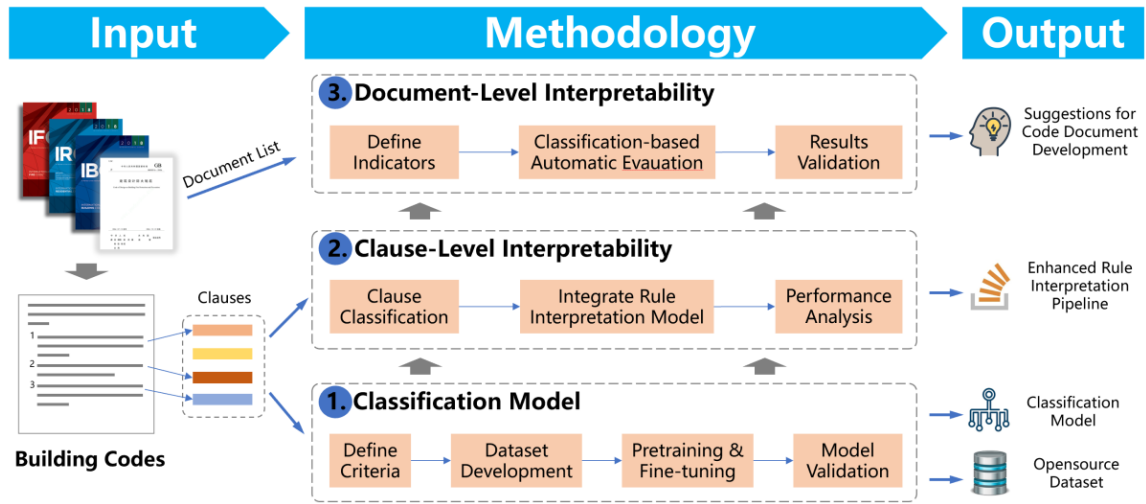

Fig. 1 Workflow of this research

## 3.1 Clause-level classification categories

The proposed clause-level category criteria aim to determine the interpretability of clauses in the design stage. Therefore, the clauses related to the construction and maintenance stages, which may require detailed and complex information from the construction site, are eliminated. Then, if the data required by the clauses can be extracted via the BIM model structure, the clauses are more likely to be interpreted. In addition, the structure of the BIM model can be simple (e.g., when data can be directly read) or complex (e.g., when geometry should be considered). Considering the above points, after studying the Chinese design codes and the data structures of BIM models, we find it useful to distinguish seven general categories of clauses, which are described in more detail below. The description includes explanations of the definitions of the clauses, the interpretability of the clauses, and typical examples of the clauses. The categories are summarized in Table 3.

### 3.1.1 Category 1: Direct

Definition: This class of clauses checks the explicit attributes and entity references that exist inside the given BIM dataset. The information is explicitly available from the model, either directly from the entities or from its associated properties with other entities (via the explicit relationship entities).

Interpretability: The data required for this type of clause can be obtained directly, so this is the simplest and the most interpretable type of clause.

Example: "The height of the enclosure walls should not be less than 2 m" (厂区周围宜设围墙，其高度不宜小于2m。) (CJJ 64-2009). The information in this clause involves the entity "wall" and the attribute "height", which can be directly obtained through the BIM model.

### 3.1.2 Category 2: Indirect

Definition: The information required for this type of clause cannot be directly extracted from the BIM model. A set of derivations and calculations should be performed on the information directly obtained from the BIM model to prepare the required information for this type of clause. Then, the check can be conducted by comparing the derived data with the data specified in the clauses.

Interpretability: The data required for this type of clause are not explicitly stored as BIM data. Additional derivations and calculations are required. However, all the data are stored in the BIM model. Therefore, this type of clause has a high degree of interpretability.

Example: "The distance between the safety exits should not exceed 120 m." (电缆隧道的安全出口间距不应超过 120m。) (GB 50987-2014). The attribute "distance" specified in this clause is not directly stored in the BIM model. However, the coordinates of the two safety exits can be obtained from the BIM model, and then the distance can be calculated based on the coordinates.

### 3.1.3 Category 3: Method

Definition: This type of clause specifies the methods and measures that should be used in a design. These methods and measures are not described in the other clauses of the building code. Therefore, an extended data structure or domain-specific knowledge is required to understand this type of clause.

Interpretability: An extended data structure or domain-specific knowledge is required to understand this type of clause. Therefore, this type of clause can be interpreted but is more difficult to interpret than the former two types.

Example: "Natural ventilation should be adopted for building ventilation" (建筑通风宜采用自然通风方式。) (GB/T 50824-2013). The method of "natural ventilation" specified in the design is not predefined in the BIM model. Thus, to address this type of clause, additional data structures should be implemented based on domain knowledge. If the extended data structure is not defined, this type of clause cannot be interpreted.

### 3.1.4 Category 4: Reference

Definition: External information should be introduced to supplement the content of this type of clause. Such external information includes pictures, formulas, tables, and other clauses or appendices in the current building code and other building codes.

Interpretability: External information is required to understand this type of clause. If the corresponding information can be found, it is difficult to interpret this type of clause. Therefore, this type of clause can be interpreted but is more difficult to interpret than the clauses contained in the direct and indirect classes.

Example: "The physical properties of steel should satisfy the value provided in Table 3.2.7." (钢材的物理性能指标应按表 3．2．7 采用。) (GB 50917-2013). The external information in "Table 3.2.7" is required to interpret this clause.

### 3.1.5 Category 5: General

Definition: This type of clause provides macro guidance for the design process. Such clauses cannot currently be handled by the BIM model.

Interpretability: This type of clause is very macroscopic and cannot be interpreted.

Example: "The materials, functions, and quality of doors and windows should meet the requirements for use." (门窗的材料、功能和质量等应满足使用要求。) (GB 50352-2019). This clause provides general design guidance, but it cannot be interpreted by the BIM model.

### 3.1.6 Category 6: Term

Definition: This type of clause defines the terms used in the codes.

Interpretability: This type of clause cannot be interpreted.

Example: "Water consumption: the amount of water consumed by users." (用水量：用户所消耗的水量。) (GB 50013-2006).

### 3.1.7 Category 7: Other

Definition: The clauses that do not belong to the above six categories are "others". This type of clause is usually a construction or maintenance process requirement, which is irrelevant to the BIM design and checking processes.

Interpretability: These clauses are difficult to interpret.

Example: "The amount of underground water for firefighting and sprinkling reserves should be replenished in time."(井下消防及洒水储备水量应能及时得到补充。) (GB 50383-2006).

Table 3 Categories of the clauses

| Category | Definition | Interpretability |
| --- | --- | --- |
| direct | The required information is explicitly available from the BIM model. | Easy |
| indirect | The required information is implicitly stored in the BIM model. A set of derivations and calculations should be performed. | Easy |
| method | An extended data structure and domain-specific knowledge are required. | Medium |
| reference | The external information, including pictures, formulas, tables, and other clauses or appendices in the current building code or other building codes, is required. | Medium |
| general | The clauses provide macro design guidance. | Hard |
| term | The clauses define the terms used in the codes. | Hard |
| other | The clauses do not belong to the above six categories. | Hard |

## 3.2 Classification dataset development

The dataset development procedure consists of three steps, including step 1: data acquisition and cleaning, step 2: data labeling, and step 3: data augmentation. The developed dataset can be found at https://github.com/SkydustZ/Text-Classification-Based-Approach-for-Evaluating-and-Enhancing-Machine-Interpretability-of-Building/tree/main/CivilRules/dataset.

Step 1: Data acquisition and cleaning. All the regulatory texts are crawled from a website containing Chinese codes (Soujianzhu, 2021) by using Python scripts. A total of 396 codes involving various disciplines are crawled, with a total of 560,509 lines. The obsolete codes are also considered because the aim of this work is rule classification, and the timeliness of the contents of the clauses have no effect on this goal. The crawled raw texts contain much redundant information, such as the names and departments of the editors, which is useless for the text classification task. In addition, the raw texts also contain some disorderly tables and redundant, garbled codes and symbols. Therefore, data cleaning is performed to filter irrelevant information, and only the regulatory texts are extracted and saved. In addition, some clauses are composed of multiple sentences that are located in separate lines. To ensure the semantic integrity of the clauses, we combine such sentences into one line. The cleaned texts have a total of 126,433 lines.

Step 2: Data labeling. After performing data cleaning, data labeling is conducted according to the clause-level category criteria proposed in Section 3.1. Domain experts manually label approximately 200 sentences for each of the 7 categories, as shown in the blue columns in Fig. 2. The dataset contains a total of 1350 sentences.

Step 3: Data augmentation. Among the 7 clause-level categories, the numbers of clauses in the direct and indirect categories are less than those of the other categories. The dataset is imbalanced, which is not conducive to model training. Note that this work focuses on the classification characteristics of the clauses, and the correctness of the content of the clauses has no effect on this task. Therefore, data augmentation can be carried out on these two categories of clauses to expand and balance the dataset. In this work, the data augmentation process involves the word replacement method, i.e., replacing the numerical values and comparison operators (e.g., more than, less than, and equal to) in one clause to generate a new clause. For example, we can replace the value "150" in the rule "The span of a single-story warehouse should not be greater than 150 m" with a new value "720". A new clause can be generated: "The span of a single-story warehouse should not be greater than 720 m". For another example, we can replace the comparison operator "less than" in the clause "The height of the rail should not be less than 1.1 m" with a new operator "more than". A new clause can also be generated. The numbers of clauses in the 7 categories after data augmentation are shown in the orange columns of Fig. 2. The dataset contains a total of 1,450 clauses.

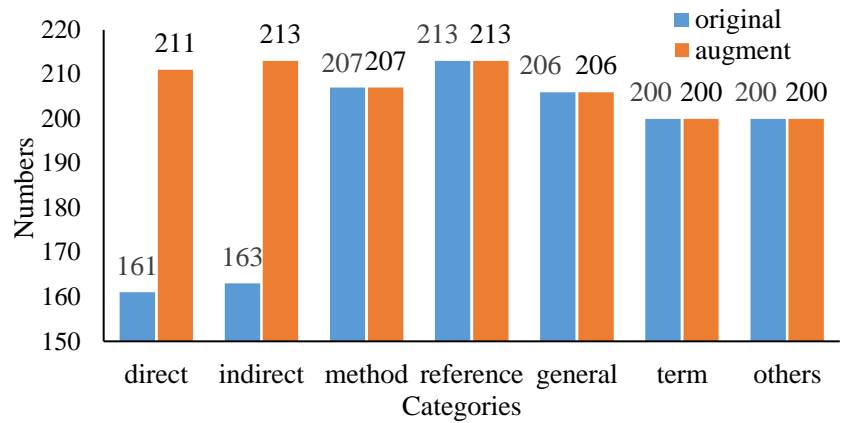

Fig. 2 Distributions of the developed datasets

## 3.3 Document-level interpretability evaluation indicator

This section illustrates the document-level interpretability evaluation indicator for a whole building code. According to clause-level interpretability, the 7 categories can be divided into three groups, as shown in Table 4. Group 1 includes the clauses in the direct and indirect categories, which are easy to interpret. Each clause in Group 1 is counted as 1 point. Group 2 includes the clauses in the method and reference categories, which can possibly be interpreted. Each clause in Group 2 is counted as 0.5 points. Group 3 includes the clauses in the remaining 3 categories, which are difficult to interpret. Each clause in Group 3 is counted as 0 points.

Table 4 Document-level interpretability evaluation indicator

|  | Group 1 | Group 2 | Group 3 |
|---|---|---|---|
| Categories | direct, indirect | method, reference | general, other, term |
| Interpretability | Easy | Medium | Hard |
| Score of each sentence | 1 | 0.5 | 0 |

Then, the average interpretability of each building code can be calculated by using equations (1) and (2).

$$score_{code} = \sum_{i=1}^{n} score_{clause_i} \qquad (1)$$

$$interpretability_{code} = \frac{score_{code}}{clause\ number_{code}} \times 100\ \% \qquad (2)$$

where $score_{code}$ is the total score of each building code. $score_{clause_i}$ is the score of $clause_i$. $clause\ number_{code}$ is the number of clauses contained in the building code.

### 3.4 Text classification based on a domain-specific pretrained model

Compared to the text classification task for subjects or topics, the classification task for interpretability is more complex and is dependent on a comprehensive understanding of sentence semantics. With the development of DL techniques and open datasets for model training, it is possible for NLP-based methods to achieve a more comprehensive understanding of regulatory texts (Fuchs, 2021). Therefore, the well-known pretrained BERT model is adopted in this work. In addition, to improve the model's performance in cases with small samples, this work adopts a domain corpus to further enhance the model; this requires no additional manual labeling efforts.

The overall workflow is shown in Fig. 3. First, the domain corpus is used to further pretrain the BERT model, and then a domain-specific BERT model is obtained (Section 3.4.1). Subsequently, the constructed text classification dataset is used to train the model to obtain the target well-trained BERT model (Section 3.4.2) that can be used for text classification prediction.

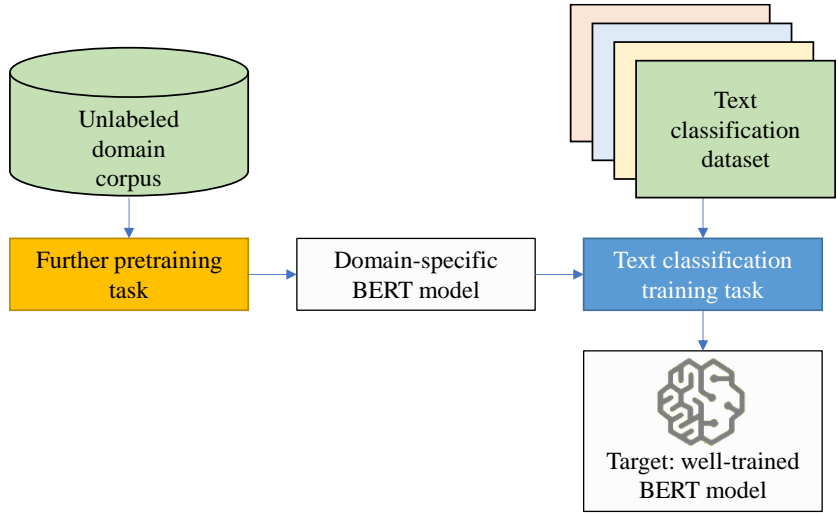

Fig. 3 The overall workflow of the adopted training strategy

### 3.4.1 Further domain corpus-enhanced pretraining

The BERT model is pretrained on the general-domain corpus (Sun et al., 2019), which has a different data distribution from that of the target domain. When the labeled dataset is very limited, the BERT model pretrained on the general-domain corpus still has difficulty obtaining satisfactory results. Therefore, to improve the performance of the pretrained model, this work adopts the domain corpus enhancement method with the domain corpus constructed by Zheng et al. (Zheng et al., 2022). We further pretrain BERT with a masked language model and subsequently complete sentence prediction tasks (Devlin et al., 2018) on the domain corpus.

The process of the adopted masked language model in the additional pretraining task is shown in Fig. 4. First, some words in the input sentence are randomly masked. The masked sentences are then used as inputs for the BERT model, while the ground truth contains the masked words. Then, the weights of the parameters in the original BERT model (i.e., the yellow part in Fig. 4) are optimized by predicting the masked words. The domain-specific BERT model can be obtained after finishing the additional pretraining task. It should be noted that the masked language model and next sentence prediction tasks are typical unsupervised tasks that do not require additional manual labeling efforts.

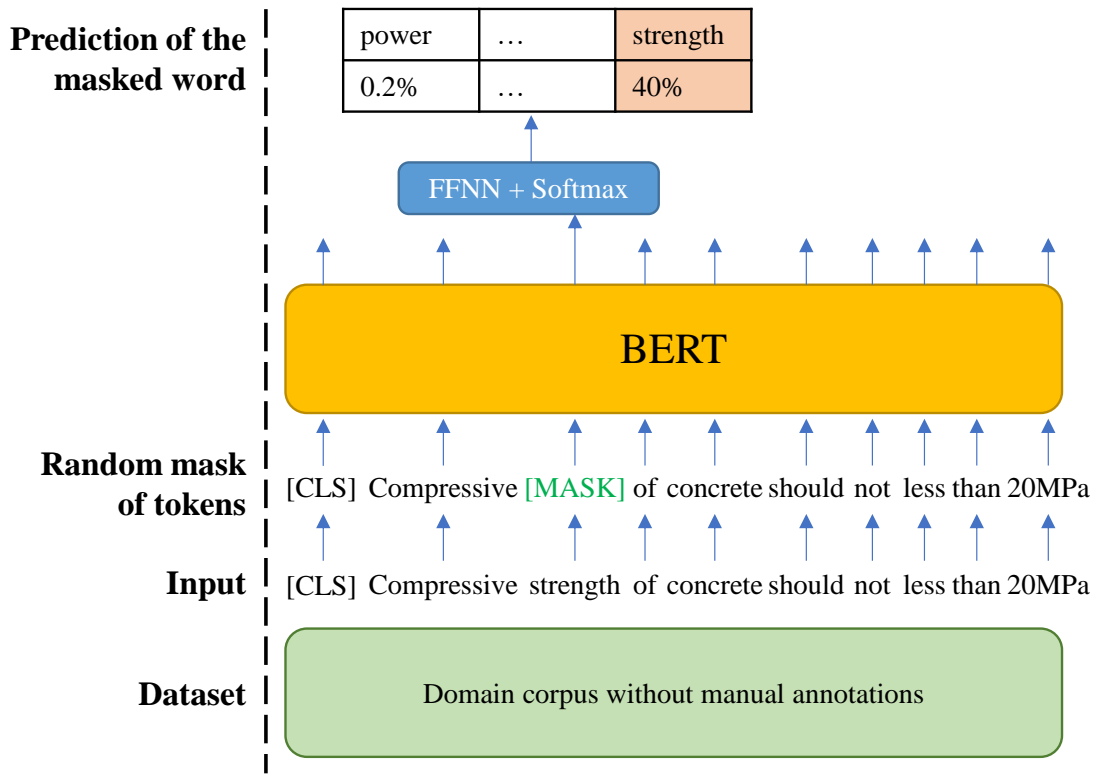

Fig. 4 Schematic of the additional pretraining task

### 3.4.2 Fine-tuning for the text classification task

After completing the additional pretraining tasks, the domain-specific BERT model is trained for the text classification task. The basic training process of the BERT-based text classification model is shown in Fig. 5. First, all tokens of each input sentence are embedded by using the pretrained domain-specific BERT model. Then, all the embeddings are encoded into contextual representations by the BERT model. Subsequently, the contextual representations are input into the fine-tuning layers to obtain the text classification prediction result. In this training process, the input of the model is a sentence, and the ground truth is a manual label.

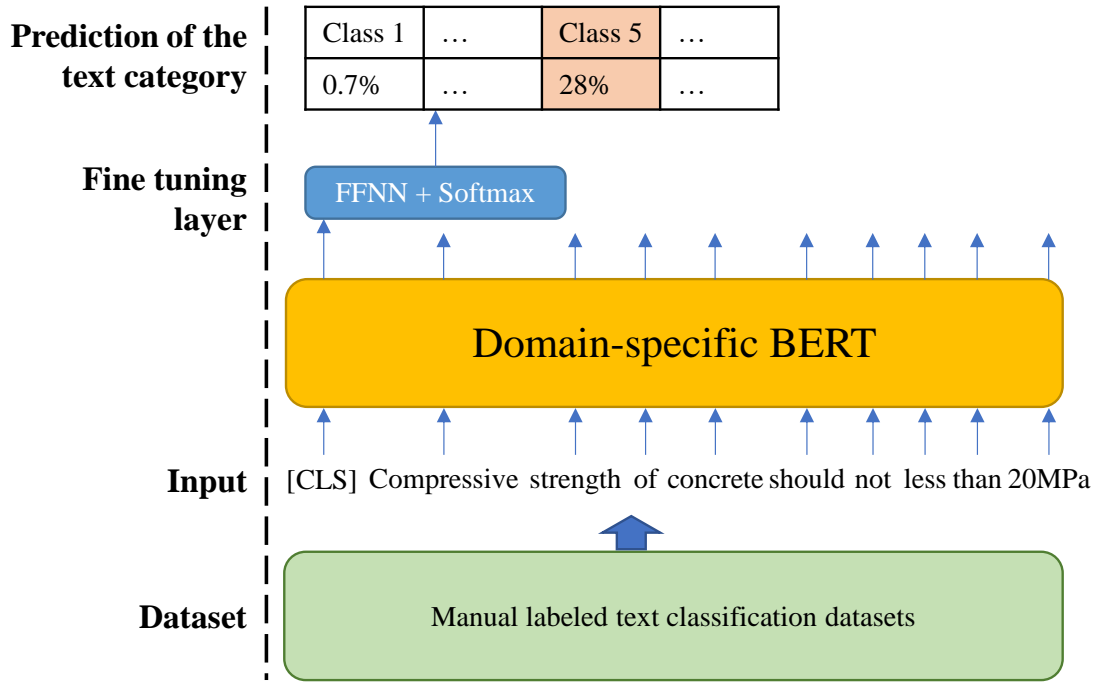

Fig. 5 Schematic of the training task for text classification

# 4 Experiments and results

## 4.1 Model pretraining and fine-tuning

For the additional pretraining task of the BERT model, an in-domain corpus, which contains 126,433 lines of Chinese regulatory texts and a total of 10,895,634 Chinese characters (Zheng et al., 2022), is adopted. The widely used bert-base-chinese model (Hugging Face, 2019) is chosen as the original BERT model. The bert-base-chinese model is the most common BERT-based model for completing tasks in Chinese; it is pretrained based on the Chinese Wikipedia corpora. Further pretraining tasks, including masked language model construction and the next sentence prediction task, are adopted (Hugging Face, 2019) and are also illustrated in Section 3.4.1. As recommended by Zheng et al. (Zheng et al., 2022), the models are further pretrained with a learning rate of 5e-5 and a batch size of 4. The further pretrained model is denoted as RuleBERT.

For the training task, which is illustrated in Section 3.4.2, the balanced text classification dataset obtained after augmentation is adopted. The dataset is randomly split into training, validation, and test datasets at a 0.8: 0.1: 0.1 ratio, where the training dataset is used to train and update the DNN model, and the validation dataset is used to test the performance of the model and chose the best hyperparameters and the corresponding best model, and the test dataset is used for the final performance evaluation.

The BERT-based models (i.e., the original BERT models and RuleBERT) are fine-tuned on the training datasets with 100 epochs and a padding size of 64. In addition, the effects of different learning rates are considered via grid searches with learning rates of 0.00007, 0.00005, 0.00003, and 0.00001.

Four types of widely used traditional DL-based text classification approaches are compared with our model. These models include (1) TextCNN (Chen, 2015), (2) TextRNN (Liu et al., 2016), (3) TextRNN with attention (TextRNN-Att for short), and (4) Transformers (Vaswani et al., 2017). The initial parameters of the word embedding representation layers of the above models are pretrained on the Chinese Wikipedia corpora (Wikipedia, 2021) by using a skip-gram model (Mikolov et al., 2013). The above DL models are then trained for 100 epochs with a padding size of 64. In addition, the effects of different learning rates are considered via grid searches with learning rates of 0.001, 0.0005, 0.00025, and 0.0001. Note that the BERT-based models have many more parameters than the traditional DL-based models, so their learning rates are different. The performance of these models on the test dataset is shown in Section 4.3.

## 4.2 Evaluation performance metrics

To measure the results, the model predictions are compared with the gold standard, and the widely used weighted average F1-score (weighted F1) is selected as the metric.

First, the precision (P), recall (R), and F1-score (F1) are calculated for each semantic label:

$$P = N_{correct}/N_{labeled} \quad (3)$$

$$R = N_{correct}/N_{true} \quad (4)$$

$$F_1 = 2PR/(P+R) \quad (5)$$

where $N_{\{correct,labeled,true\}}$ denotes the number of {correctly labeled by model, labeled by model, true} elements for a label. Finally, the weighted average F1-score is calculated to represent the overall performance ($n_i$ denotes the number of elements in the $i$-th semantic label):

$$\text{Weighted } F_1 = \left(\sum_i n_i F_{1,i}\right) / \sum_i n_i \quad (6)$$

## 4.3 Performance evaluation

The performance of the well-trained text classification models is shown in Fig. 6, and the following conclusions are drawn. First, the original BERT model performs better than the other traditional DL models, which is in line with the results of Fang et al. (Fang et al., 2020). Second, after completing the additional pretraining task on the in-domain corpus, the enhanced BERT model (i.e., RuleBERT) performs even better than the original BERT model. RuleBERT achieves the global best weighted F1-score of 93.60%, which corresponds to a state-of-the-art effect. Note that the further pretraining task utilizes the unlabeled in-domain corpus, which does not require manual annotation effort.

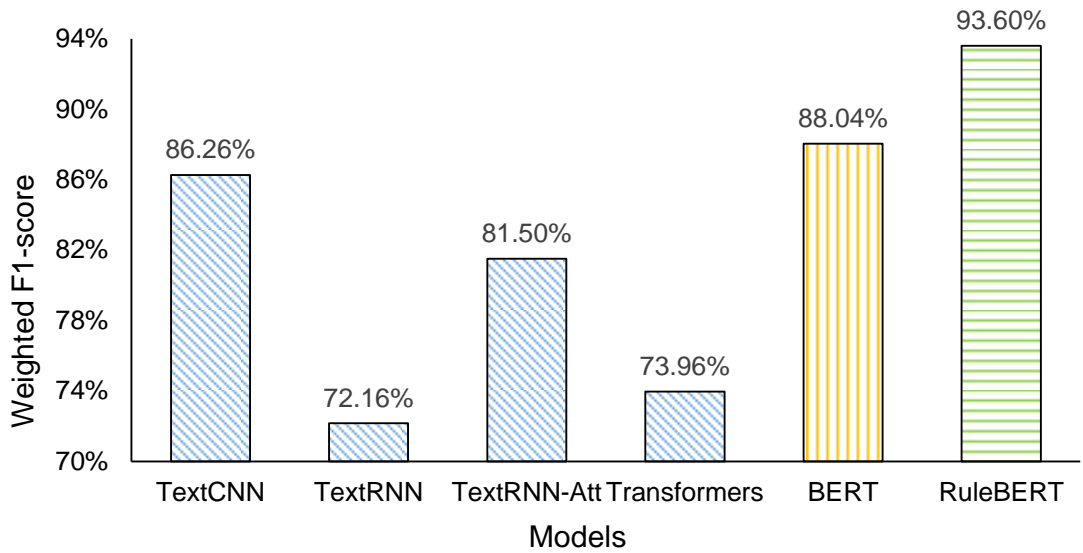

Fig. 6 Weighted F1-scores achieved on the text classification datasets

## 5 Quantitative analysis of rule interpretation improvement

Before automated rule interpretation, our text classification-based interpretability evaluation method can filter those clauses with low machine interpretability. Then the clauses with high machine interpretability can be interpreted using an automated rule interpretation algorithm. In this way, the error rate of the automated rule interpretation algorithm can be reduced.

Therefore, we carried out an experiment to demonstrate the improvement achieved by the proposed interpretability evaluation method in terms of rule interpretation. The successful and correct interpretation rates before text classification are taken as the benchmark, and the percentages of clauses successfully and correctly interpreted before and after the classification process are compared. In order to evaluate the accuracy of the automated rule interpretation results, the correctness of the automated interpretation results is determined in a sentence-by-sentence manner by domain experts.

Specifically, this experiment includes three steps. Step 1: Before conducting text classification, the automated rule interpretation method is used to interpret all the clauses derived from one regulatory document. Then, domain experts manually check the automated interpreted results. Subsequently, the percentage of clauses that are successfully and correctly interpreted is calculated. Step 2: The proposed text classification-based interpretability evaluation method is executed. The clauses that are difficult to interpret (i.e., the clauses in the general category, the "other" category, and the terms category) are excluded. Step 3: After classification, the automated rule interpretation method is applied to the remaining clauses. Then, the percentage of clauses successfully and correctly interpreted after text classification is calculated.

In this work, the regulatory texts obtained are from the *Chinese Code for Fire Protection Design of Buildings* (GB 50016-2014). After performing sentence splitting, the specification contains 1004 lines of clauses. The automated interpretation method proposed by Zhou et al. (Zhou et al., 2022; Zheng et al., 2022) is adopted to perform rule interpretation. The adopted automated interpretation method includes three main steps. First, a syntax tree structure with seven semantic elements (i.e., obj, sobj, prop, cmp, Rprop, ARprop, and Robj) is proposed to represent the roles and relations of concepts in regulatory text. Second, a deep learning model with the transfer learning technique is utilized to label the semantic elements in a sentence. Finally, a set of CFGs is built to parse a labeled sentence into the language-independent tree structure, from which computable checking rules can be generated. It should be noted that although the experiment in this work adopted the automated interpretation algorithm proposed by Zhou et al. (Zhou et al., 2022), alternative automated rule interpretation algorithms can also be utilized in other systems. The well-trained RuleBERT (Section 4.3) is utilized to perform text classification.

Before the text classification operation, a total of 1004 sentences are interpreted, and the number of clauses correctly and successfully interpreted is shown in Table 6. The results of text classification are shown in Table 5. The clauses that are not suitable for interpretation are excluded, and the remaining 749 clauses are suitable for interpretation. It can be seen from Table 6 that after performing text classification, the accuracy of the automated interpretation results is improved from 68% to 72%. The experimental results also validate the reliability and validity of the proposed method.

Table 5 Rule classification results

| Total | direct | indirect | reference | method | general | others | term |
|---|---|---|---|---|---|---|---|
| 1004 | 280 | 193 | 119 | 157 | 70 | 112 | 73 |

Table 6 Rule interpretation results

|  | Before text classification | After text classification |
|---|---|---|
| Number of input clauses | 1004 | 749 |
| Number of successful clauses | 679 | 539 |
| Percentage of successful clauses | 68% | 72% |

# 6 Interpretability evaluation for different Chinese building codes
## 6.1 Data collection for building codes
### 6.1.1 Fundamental building codes in different domains and levels

The proposed text classification-based interpretability evaluation method is first used to assess and analyze Chinese fundamental building codes from different domains. To perform the evaluation, 26 codes from 8 building design subdomains are collected first. The domains include structural, water supply and drainage, electrical, heating and ventilation, intelligence, fire protection, energy efficiency and environmental protection, and architecture domains. The numbers of building codes and clauses in different domains are shown in Table 7. The codes are divided into two groups, GB and HB, based on their levels. GB stands for national standards, and HB stands for industrial standards (e.g., JGJ, CJJ, and CECS).

Table 7 Chinese fundamental building codes in different domains and levels

| Domains | Number of building codes in GB level | Number of clauses | Number of building codes in HB level | Number of clauses |
|---|---|---|---|---|
| Structural | 4 | 2864 | 1 | 1032 |
| Water supply and drainage | 2 | 1643 | 1 | 146 |
| Electrical | 1 | 259 | 2 | 3310 |
| Heating and ventilation | 1 | 985 | 2 | 703 |
| Intelligence | 1 | 530 | 1 | 254 |
| Fire protection | 3 | 1192 | 1 | 86 |
| Energy efficiency and environmental protection | 1 | 311 | 1 | 205 |
| Architecture | 2 | 625 | 2 | 353 |
| Total | 15 | 8409 | 11 | 6089 |
| All | | | 26 | 14498 |

**6.1.2 Building codes from the fire protection domain**

The proposed text classification-based interpretability evaluation method is then used to assess and analyze the codes at different levels of the fire protection domain. A total of 134 building codes are collected the fire protection domain at the national level (GB), the industrial level (HB), and the local level (DB). As the numbers of clauses in some codes are too small, only the codes with more than 50 clauses are retained for further analysis (97 total). Among the 97 codes, 62 are at the GB level, 22 are at the HB level, and 13 are at the DB level.

**6.2 Evaluation of fundamental building codes from different domains**

After performing data collection, the well-finetuned RuleBERT model described in Section 4 is used to classify all the clauses into the 7 categories defined in Section 3. Then, the quantitative evaluation method developed in Section 3.3 is used to assess the document-level interpretability obtained from different domains and at different levels.

The scores for the different domains and levels of the fundamental building codes are shown in Table 8. For all the fundamental codes, the overall average interpretability is 34.40%. Other researchers manually evaluate the interpretability of Chinese fundamental codes and they found that the average interpretability of Chinese fundamental codes is 34% (Gu, 2021). Our automated evaluation result matches well with the result evaluated by professors manually, which validates the reasonability of our methods. Furthermore, the average interpretability of all codes at the GB level is 35.51%, which is higher than the 32.82% achieved for all codes at the HB level. This indicates that the interpretability of the codes at the GB level is better than that of codes at the HB level.

Regarding the codes in different domains, the fire protection domain achieves the highest interpretability, with 58.64% interpretability at the GB level and 53.49% at the HB level. This means that more than half of the clauses in the fire protection domain have the potential to be automatically interpreted. The average interpretability of the architecture domain is the second-highest value, exceeding 40%. The interpretability scores of the intelligence domain and the energy efficiency and environmental protection domain are lower than 20%.

Table 8 Interpretability of building codes in different domains

| Domains | The national level (GB) | The industrial level (HB) |
|---|---|---|
| Structural | 34.27% | 37.12% |
| Water supply and drainage | 35.85% | 30.81% |
| Electrical | 29.93% | 32.69% |
| Heating and ventilation | 26.29% | 31.65% |
| Intelligence | 10.76% | 10.63% |
| Fire protection | 58.64% | 53.49% |
| Energy efficiency and environmental protection | 16.88% | 27.81% |
| Architecture | 43.84% | 38.38% |
| Total | 35.51% | 32.82% |
| All | | 34.40% |

## 6.3 Evaluation of the building codes from the fire protection domain

Then the proposed quantitative evaluation method developed in Section 3.3 is used to assess the document-level interpretability of the building codes from the fire protection domain. The interpretability values of codes obtained from different levels are shown in Table 9. The average interpretability values show that the interpretability ranking is GB > HB > DB, which is consistent with the results in Section 6.2, i.e., GB > HB.

When the interpretability value of one building code is greater than 50%, more than half of its clauses can be interpreted. Therefore, these building codes are defined as highly interpretable codes. An analysis of the highly interpretable codes at different levels is conducted. The results show that over 24.19% of the codes at the GB level are highly interpretable, 9.10% of the codes at the HB level are highly interpretable, and 7.69% of the codes at the DB level are highly interpretable.

The above analysis shows that for the Chinese codes obtained from the fire protection domain at different levels, the interpretability ranking is GB > HB > DB.

Table 9 The interpretability of the codes in the fire protection domain

| Levels | Max interpretability value | Average interpretability value | Highly interpretable code |
|---|---|---|---|
| GB | 64.49% | 36.40% | 24.19% |
| HB | 52.63% | 29.41% | 9.10% |
| DB | 60.32% | 26.38% | 7.69% |

## 7 Discussion

This work contributes to the overall body of knowledge on four main levels.

(1) This work proposes novel clause-level category criteria for rule interpretation based on an investigation of Chinese design codes and the data structures of BIM models. The proposed category criteria can evaluate the complexity and interpretability of clauses and building codes. Based on the category criteria, regulatory texts were collected, and a text classification dataset was constructed for DL model training via manual annotation. A data augmentation method was proposed to solve the imbalance problem faced by this dataset. The constructed dataset can be used to train text classification models for clauses in the civil domain and can also be used to test the performance of different text classification methods. The dataset has been made open source, so it can contribute to the development of ARC.

(2) We adopted advanced NLP and a pretrained DL model to automate the text classification-based interpretability evaluation process. To address the small sample issue, we utilized an additional pretraining method with the widely used original BERT model and obtained the RuleBERT model. We compared and evaluated the performance of our approach against other DL-based text classification approaches. The experimental results demonstrated that our further pretrained DL model (RuleBERT) could achieve the best text classification performance. Note that the additional pretraining method is an unsupervised task. RuleBERT performed better than the original BERT model without additional manual labeling efforts. The training codes and well-trained models are open-sourced and could be found at https://github.com/SkydustZ/Text-Classification-Based-Approach-for-Evaluating-and-Enhancing-Machine-Interpretability-of-Building.

(3) In addition, we carried out an experiment to demonstrate the effect of our automated text classification method in terms of automated rule interpretation. The results showed that when utilizing our method, the success rate of automated rule interpretation was improved by 4%.

(4) A document-level interpretability evaluation indicator was proposed based on the clause-level category criteria. More than 150 Chinese design codes were collected for evaluation purposes. Subsequently, the well-trained RuleBERT associated with the proposed document-level interpretability evaluation indicator was utilized to complete two evaluation tasks. First, the interpretability of Chinese fundamental codes in different domains and at different levels was evaluated. For all the fundamental codes, the overall average interpretability was 34.40%. The building codes in the fire protection domain were most interpretable, with an interpretability score exceeding 50%. Second, the interpretability of 97 building codes in the fire protection domain was evaluated. The analysis showed that for the Chinese codes of the fire protection domain at different levels, the interpretability ranking was GB (the national level) > HB (the industrial level) > DB (the local level). To the best of our knowledge, this is the first effort on interpretability evaluation for large-scale Chinese building codes. The conducted interpretability evaluation can provide guidance for the optimization and revision of the existing codes. For example, to achieve better interpretability, the codes in other domains can refer to the codes in the fire protection domain, and the codes at other levels can refer to the codes at the GB level. In addition, the proposed evaluation method is conducive to the development of automated rule interpretation methods. For example, future work can focus more on clauses that have the potential to be automatically interpreted.

According to the interpretability analysis conducted on the fundamental regulatory documents, approximately one-third of the existing clauses can be automatically interpreted, which means that there is still a long way to go to achieve full ARC. In other words, to realize full ARC for all the existing building codes, relying solely on the intelligence of algorithms and machines may not be sufficient. Human–machine intelligence is expected to promote ARC development in the following ways. For example, first, domain experts can revise the existing regulatory documents to (1) simplify the structures of the documents and (2) improve consistency and clarity regarding language usage. In this way, the interpretability of regulatory documents may increase. Second, a domain-specific knowledge database should be established (Wu et al., 2022b; Zhou et al., 2021); this would require domain experts' efforts and is essential for a machine to understand the meanings of clauses. As for the policymakers or managers, they might need to rethink the code-writing process and invite computer experts to each code development committee. Experts in civil engineering make up the majority of the present code development committee. Today, the codes are meant to be read by engineers (human beings). These codes written by the present code development committee may have low interpretability because experts in civil engineering may have a limited understanding of computer technologies. Managers may think about integrating computer experts into the code development committee to make the committee has both civil engineering and computer science insights. Computer experts can also develop some interpretability evaluation tools to assist the code-writing process.

Despite the success of our proposed rule classification criteria and model, the following limitations need to be acknowledged. First, the proposed criteria are still a preliminary attempt in terms of interpretability evaluation. Therefore, more detailed rule classification criteria can be developed in the future. For example, future classification criteria can consider rule interpretation methods. Thus, the most proper predefined interpretation methods can be assigned to the corresponding clauses to interpret more complex clauses. Second, the reasons why the clauses are difficult to interpret can be further analyzed and manually highlighted. Thus, the optimization and revision methods for the existing codes can be summarized to achieve better automated rule interpretation. Third, the developed datasets can be enriched to further improve the performance of DL models. Fourth, for the small sample issue, despite the use of our pretraining method, other methods should be further explored. For example, prompt learning has been proposed in recent years and is designed for few-shot or zero-shot learning. The last point is that future research can be based on the visualization of attention patterns in BERT models to understand what is going on inside the "black box" of BERT models and increase the explainability of BERT models (Tagarelli & Simeri, 2022). This may increase the understanding of the complex patterns and relationships of building codes.

## 8 Conclusion

In this research, we first proposed clause-level text classification category criteria that can evaluate the interpretability of clauses. Based on the proposed category criteria, a dataset was constructed for DL model training. A data augmentation method was proposed to solve the imbalance problem faced by the dataset. Second, we developed an automated text classification method utilizing an advanced pretrained DL model and transfer learning techniques. Our further pretrained RuleBERT model achieved the best performance and outperforms the existing CNN- or RNN-based methods, improving the F1-score from 72.16% to 93.60%. Third, the proposed automated text classification method can enhance downstream Automated rule interpretation methods with an improvement of 4%. Fourth, we proposed corresponding indicator and method to evaluate the document-level interpretability of building codes based on the automated text classification method and then we evaluated more than 150 Chinese design codes from different domains and at different levels. The analysis showed that (1) approximately 34.40% of clauses in Chinese fundamental codes have the potential to be automatically interpreted; (2) among different domains, the building codes in the fire protection domain were most interpretable, with an interpretability score exceeding 50%; and (3) for the codes of the fire protection domain at different levels, the interpretability ranking was GB (the national level) > HB (the industrial level) > DB (the local level). To the best of our knowledge, this is the first effort on interpretability evaluation for large-scale Chinese building codes.

This research provides an efficient approach for quickly evaluating the interpretability of clauses and building codes, which can be used in multiple ways, including improper sentences filtering in ARC and interpretability evaluation for large-scale codes. This research also published the labeled text classification dataset for future exploration, validation, and benchmarking in the ARC field. In addition, the analysis in this work shows that to achieve fully automated rule checking, cooperation between domain experts and algorithms is required (i.e., human–machine intelligence). These contributions can significantly promote the research and application of ARC.

Further research is required to expand the category criteria and datasets for more detailed evaluations and predefined interpretation method recommendations. Regarding the small sample issue, despite the success of our pretrained model, more techniques should be explored. For example, the prompt learning method, which is designed for few or zero-shot learning, may be applicable.

## Acknowledgments

The authors are grateful for the financial support received from the National Natural Science Foundation of China (no. 51908323 and no. 72091512), the National Key R&D Program (no. 2019YFE0112800), and the Tencent Foundation through the XPLORER PRIZE.

# References：

[1] Wu, C., Li, X., Guo, Y., Wang, J., Ren, Z., Wang, M., & Yang, Z. (2022a). Natural language processing for smart construction: Current status and future directions. Automation in Construction, 134, 104059. https://doi.org/10.1016/j.autcon.2021.104059

[2] Liao, W., Lu, X., Huang, Y., Zheng, Z., & Lin, Y. (2021). Automated structural design of shear wall residential buildings using generative adversarial networks. Automation in Construction, 132, 103931. https://doi.org/10.1016/j.autcon.2021.103931

[3] Ismail, A. S., Ali, K. N., & Iahad, N. A. (2017). A review on BIM-based automated code compliance checking system. In 2017 International Conference on Research and Innovation in Information Systems (ICRIIS), 1-6. https://doi.org/10.1109/ICRIIS.2017.8002486

[4] Eastman, C., Lee, J. M., Jeong, Y. S., & Lee, J. K. (2009). Automatic rule-based checking of building designs. Automation in Construction, 18(8), 1011-1033. https://doi.org/10.1016/j.autcon.2009.07.002

[5] Sobhkhiz, S., Zhou, Y. C., Lin, J. R., & El-Diraby, T. E. (2021). Framing and Evaluating the Best Practices of IFC-Based Automated Rule Checking: A Case Study. Buildings, 11(10), 456. https://doi.org/10.3390/buildings11100456

[6] Fuchs, S. (2021). Natural language processing for building code interpretation: systematic literature review report.

[7] Zhang, R., & El-Gohary, N. (2021a). A deep neural network-based method for deep information extraction using transfer learning strategies to support automated compliance checking. Automation in Construction, 132, 103834. https://doi.org/10.1016/j.autcon.2021.103834

[8] Zhou, Y. C., Zheng, Z., Lin, J. R., & Lu, X. Z. (2022). Integrating NLP and context-free grammar for complex rule interpretation towards automated compliance checking. Computers in Industry, 142, 103746. https://doi.org/10.1016/j.compind.2022.103746

[9] Soliman-Junior, J., Tzortzopoulos, P., Baldauf, J. P., Pedo, B., Kagioglou, M., Formoso, C. T., & Humphreys, J. (2021). Automated compliance checking in healthcare building design. Automation in Construction, 129, 103822. https://doi.org/10.1016/j.autcon.2021.103822

[10] Zhou, P., & El-Gohary, N. (2016). Domain-specific hierarchical text classification for supporting automated environmental compliance checking. Journal of Computing in Civil Engineering, 30(4), 04015057. https://doi.org/10.1061/(ASCE)CP.1943-5487.0000513

[11] Song, J., Kim, J., & Lee, J. K. (2018). NLP and deep learning-based analysis of building regulations to support automated rule checking system. In ISARC. Proceedings of the International Symposium on Automation and Robotics in Construction (Vol. 35, pp. 1-7). IAARC Publications. https://www.proquest.com/docview/2123611147?pq-origsite=gscholar&fromopenview=true

[12] Zhang, J., & El-Gohary, N. M. (2015). Automated information transformation for automated regulatory compliance checking in construction. Journal of Computing in Civil Engineering, 29(4), B4015001. https://doi.org/10.1061/(ASCE)CP.1943-5487.0000427

[13] Zhang, R., & El-Gohary, N. (2021b). Clustering-Based Approach for Building Code Computability Analysis. Journal of Computing in Civil Engineering, 35(6), 04021021. https://doi.org/10.1061/(ASCE)CP.1943-5487.0000967

[14] Solihin, W., & Eastman, C. (2015). Classification of rules for automated BIM rule checking development. Automation in construction, 53, 69-82. https://doi.org/10.1016/j.autcon.2015.03.003

[15] Zhang, J., & El-Gohary, N. M. (2016). Semantic NLP-based information extraction from construction regulatory documents for automated compliance checking. Journal of Computing in Civil Engineering, 30(2), 04015014. https://doi.org/10.1061/(ASCE)CP.1943-5487.0000346

[16] Salama, D. M., & El-Gohary, N. M. (2016). Semantic text classification for supporting automated compliance checking in construction. Journal of Computing in Civil Engineering, 30(1), 04014106. https://doi.org/10.1061/(ASCE)CP.1943-5487.0000301

[17] Uhm, M., Lee, G., Park, Y., Kim, S., Jung, J., & Lee, J. K. (2015). Requirements for computational rule checking of requests for proposals (RFPs) for building designs in South Korea. Advanced Engineering Informatics, 29(3), 602-615. https://doi.org/10.1016/j.aei.2015.05.006

[18] Malsane, S., Matthews, J., Lockley, S., Love, P. E., & Greenwood, D. (2015). Development of an object model for automated compliance checking. Automation in Construction, 49, 51-58. https://doi.org/10.1016/j.autcon.2014.10.004

[19] Manning, C., & Schutze, H. (1999). Foundations of statistical natural language processing. MIT press.

[20] Caldas, C. H., Soibelman, L., & Han, J. (2002). Automated classification of construction project documents. Journal of Computing in Civil Engineering, 16(4), 234-243. https://doi.org/10.1061/(ASCE)0887-3801(2002)16:4(234)

[21] Hassan, F. U., & Le, T. (2020). Automated requirements identification from construction contract documents using natural language processing. Journal of Legal Affairs and Dispute Resolution in Engineering and Construction, 12(2), 04520009. https://doi.org/10.1061/(ASCE)LA.1943-4170.0000379

[22] Cheng, M. Y., Kusoemo, D., & Gosno, R. A. (2020). Text mining-based construction site accident classification using hybrid supervised machine learning. Automation in Construction, 118, 103265. https://doi.org/10.1016/j.autcon.2020.103265

[23] Fang, W., Luo, H., Xu, S., Love, P. E., Lu, Z., & Ye, C. (2020). Automated text classification of near-misses from safety reports: An improved deep learning approach. Advanced Engineering Informatics, 44, 101060. https://doi.org/10.1016/j.aei.2020.101060

[24] Tian, D., Li, M., Shi, J., Shen, Y., & Han, S. (2021). On-site text classification and knowledge mining for large-scale projects construction by integrated intelligent approach. Advanced Engineering Informatics, 49, 101355. https://doi.org/10.1016/j.aei.2021.101355

[25] Zhong, B., Xing, X., Luo, H., Zhou, Q., Li, H., Rose, T., & Fang, W. (2020). Deep learning-based extraction of construction procedural constraints from construction regulations. Advanced Engineering Informatics, 43, 101003. https://doi.org/10.1016/j.aei.2019.101003

[26] Salton, G., & Buckley, C. (1988). Term-weighting approaches in automatic text retrieval. Information processing & management, 24(5), 513-523. https://doi.org/10.1016/j.autcon.2021.103822

[27] Mikolov, T., Chen, K., Corrado, G., & Dean, J. (2013). Efficient estimation of word representations in vector space. arXiv preprint arXiv:1301.3781.

[28] Liu, P., Qiu, X., & Huang, X. (2016). Recurrent neural network for text classification with multi-task learning. arXiv preprint arXiv:1605.05101.

[29] Lai, S., Xu, L., Liu, K., & Zhao, J. (2015, February). Recurrent convolutional neural networks for text classification. In Twenty-ninth AAAI conference on artificial intelligence.

[30] Johnson, R., & Zhang, T. (2017, July). Deep pyramid convolutional neural networks for text categorization. In Proceedings of the 55th Annual Meeting of the Association for Computational Linguistics (Volume 1: Long Papers) (pp. 562-570).

[31] LeCun, Y., Bengio, Y., & Hinton, G. (2015). Deep learning. nature, 521(7553), 436-444. https://doi.org/10.1038/nature14539

[32] Xu, X., & Cai, H. (2021). Ontology and rule-based natural language processing approach for interpreting textual regulations on underground utility infrastructure. Advanced Engineering Informatics, 48, 101288. https://doi.org/10.1016/j.aei.2021.101288

[33] Devlin, J., Chang, M. W., Lee, K., & Toutanova, K. (2018). Bert: Pre-training of deep bidirectional transformers for language understanding. arXiv preprint arXiv:1810.04805.

[34] Tagarelli, A., & Simeri, A. (2022). Unsupervised law article mining based on deep pre-trained language representation models with application to the Italian civil code. Artificial Intelligence and Law, 30(3), 417-473. https://doi.org/10.1007/s10506-021-09301-8

[35] Tian, D., Li, M., Han, S., & Shen, Y. (2022). A Novel and Intelligent Safety-Hazard Classification Method with Syntactic and Semantic Features for Large-Scale Construction Projects. Journal of Construction Engineering and Management, 148(10), 04022109. https://doi.org/10.1061/(ASCE)CO.1943-7862.0002382

[36] Wang, Z., Wang, L., Huang, C., Sun, S., & Luo, X. (2022). BERT-based 25hinese text classification for emergency management with a novel loss function. Applied Intelligence, 1-12. https://doi.org/10.1007/s10489-022-03946-x

[37] Sun, C., Qiu, X., Xu, Y., & Huang, X. (2019, October). How to fine-tune bert for text classification?. In China National Conference on Chinese Computational Linguistics (pp. 194-206). Springer, Cham.

[38] Zheng, Z., Lu, X. Z., Chen, K. Y., Zhou, Y. C., & Lin, J. R. (2022). Pretrained domain-specific language model for natural language processing tasks in the AEC domain. Computers in Industry, 142, 103733. https://doi.org/10.1016/j.compind.2022.103733

[39] The Ministry of Public Security of the People's Republic of China (2009). Code for design of night soil treatment plant (CJJ 64-2009), Ministry of Construction of Peoples Republic of China, Beijing, China. (in Chinese)

[40] The Ministry of Public Security of the People's Republic of China (2014). Code for fire protection design of hydraulic engineering (GB 50987-2014), Ministry of Construction of Peoples Republic of China, Beijing, China. (in Chinese)

[41] The Ministry of Public Security of the People's Republic of China (2013). Design standard for energy efficiency of rural residential buildings (GB/T 50824-2013), Ministry of Construction of Peoples Republic of China, Beijing, China. (in Chinese)

[42] The Ministry of Public Security of the People's Republic of China. (2013). Code for design of steel and concrete composite bridges (GB 50917-2013), Ministry of Construction of Peoples Republic of China, Beijing, China. (in Chinese)

[43] The Ministry of Public Security of the People's Republic of China (2006). Standard for design of outdoor water supply engineering (GB 50013-2006), Ministry of Construction of Peoples Republic of China, Beijing, China. (in Chinese)

[44] The Ministry of Public Security of the People's Republic of China (2006). Code for design of the fire protecting, sprinkling system in underground coalmine (GB 50383-2006), Ministry of Construction of Peoples Republic of China, Beijing, China. (in Chinese)

[45] The Ministry of Public Security of the People's Republic of China (2019). Uniform standard for design of civil buildings (GB 50352-2019), Ministry of Construction of Peoples Republic of China, Beijing, China. (in Chinese)

[46] Soujianzhu. Chinese Rules. https://www.soujianzhu.cn/default.aspx (accessed: June 22, 2021). (in Chinese)

[47] Hugging Face. (2019). Bert-base-chinese. https://huggingface.co/bert-base-chinese/tree/main (Access on 2021-12-11)

[48] Chen, Y. (2015). Convolutional neural network for sentence classification (Master's thesis, University of Waterloo). http://hdl.handle.net/10012/9592

[49] Liu, P., Qiu, X., & Huang, X. (2016). Recurrent neural network for text classification with multi-task learning. arXiv preprint arXiv:1605.05101.

[50] Vaswani, A., Shazeer, N., Parmar, N., Uszkoreit, J., Jones, L., Gomez, A. N., Kaiser, Ł., & Polosukhin, I. (2017).

Attention is all you need. In Advances in neural information processing systems (pp. 5998-6008).

[51] Wikipedia. (2021a). Wikimedia Downloads. https://dumps.wikimedia.org/  (Access on 2021-12-11)

[52] The Ministry of Public Security of the People's Republic of China (2014), Code for fire protection design of buildings (GB 50016-2014), Ministry of Construction of Peoples Republic of China, Beijing, China. (in Chinese)

[53] Zheng, Z., Zhou, Y. C., Lu, X. Z., & Lin, J. R. (2022). Knowledge-informed semantic alignment and rule interpretation for automated compliance checking. Automation in Construction, 142, 104524. https://doi.org/10.1016/j.autcon.2022.104524

[54] Gu, M. Make BIM from can be seen to can be trusted. In Proceedings of the 8th International Conference on BIM Technology (pp. 14). https://www.doc88.com/p-99829288787126.html (In Chinese)

[55] Wu, L.T, Lin, J.R., Leng, S., Li, J.L., Hu, Z.Z. (2022b). Rule-based Information Extraction for Mechanical-Electrical-Plumbing-Specific Semantic Web. Automation in Construction, 135, 104108. doi: 10.1016/j.autcon.2021.104108 http://doi.org/10.1016/j.autcon.2021.104108

[56] Zhou, Y.C., Lin, J.R.*, She, Z.T. (2021). Automatic Construction of Building Code Graph for Regulation Intelligence. Proceedings of the 2021 International Conference on Construction and Real Estate Management (ICCREM 2021), 248-254. Beijing, China. https://doi.org/10.1061/9780784483848.028